# A Data-Driven Reinforcement Learning Solution Framework for Optimal and Adaptive Personalization of a Hip Exoskeleton

Xikai Tu, Minhan Li, Ming Liu, Jennie Si, *Fellow, IEEE* and He (Helen) Huang, *Senior* Member, *IEEE*

*Abstract*—Robotic exoskeletons are exciting technologies for augmenting human mobility. However, designing such a device for seamless integration with the human user and to assist human movement still is a major challenge. This paper aims at developing a novel data-driven solution framework based on reinforcement learning (RL), without first modeling the human-robot dynamics, to provide optimal and adaptive personalized torque assistance for reducing human efforts during walking. Our automatic personalization solution framework includes the assistive torque profile with two control timing parameters (peak and offset timings), the least square policy iteration (LSPI) for learning the parameter tuning policy, and a cost function based on transferred work ratio. The proposed controller was successfully validated on a healthy human subject to assist unilateral hip extension in walking. The results showed that the optimal and adaptive RL controller as a new approach was feasible for tuning assistive torque profile of the hip exoskeleton that coordinated with human actions and reduced activation level of hip extensor muscle in human.

*Keywords*- Exoskeleton, reinforcement learning, optimal adaptive control, data driven

I. INTRODUCTION

Lower limb robotic exoskeletons are promising technologies to augment the locomotion performance and energetics of both healthy individuals and persons with motor deficits [1]. Studies have shown that the human hip joint plays a critical role in walking with the highest delivery of positive mechanical power during a gait cycle [2, 3]. Therefore, powered hip exoskeletons have been actively researched for potential deployment [4]-[8]. Compliance between the device and its human user is an essential factor in design considerations.

Several prototypes of hip exoskeletons have been reported, featuring compliance between the device and its human user. Kang et al [8] built a robotic hip exoskeleton with two degrees of freedom (DOFs) and a weight of around 7 kg, which was actuated by two series elastic actuators (SEAs) with a peak torque of 60 Nm. Giovacchini et al [9] developed an active 4-DOF hip exoskeleton named APO weighing about 4.2 kg. Wherein, hip flexion/extension were actively actuated by the SEAs while abduction/adduction were passive. A robotic hip exoskeleton was developed by Chen et al [10] to support lift movement assistance of heavy tasks with hip flexion and extension movements, which was powered by two SEAs that achieved a peak torque of 22 Nm. Masood et al [11] have developed a robotic exoskeleton with 12 DOFs to assist able-bodied people with manual handling tasks. Only hip flexion and extension were active with an output torque of approximately 53 Nm but a weight of 11.6 kg, which were actuated by two parallel elastic actuators (PEAs). Our previously developed a SEA-driven 4-DOF hip exoskeleton, called NREL-Exo, in which both hip flexion/extension and hip adduction/abduction were actively controlled [12]-[14]. While these devices have achieved some level of success, they are restrained by the SEA for a narrow bandwidth and limited stiffness range. Additionally, their bulkiness also limits their application. In this paper, we present a newly developed lightweight, compact, compliant, and modular 2-DOF robotic hip exoskeleton with powered flexion and extension motions.

Automatic personalization of the control of a hip exoskeleton to fit its user has been challenging. Currently, there are mainly three control schemes, which almost all use fixed control parameters. Direct regulation of force or torque is one of the most common hierarchical architecture control methods in robotic hip exoskeletons. The control system includes the high- and low-level layers in which the high level is responsible for estimating the gait phases of the wearer while the low level controls torque or force, in order to provide pre-defined assistance to the wearers [15]-[17]. Another widely used control approach is impedance or admittance control. It is a compliance-based control method that can produce the complementary portion of the torque necessary to complete a compliant movement for the wearers. The integral admittance shaping algorithm was proposed to control the robotic hip exoskeleton in the research of Nagarajan et al [18]. The adaptive oscillator-based method is the third control scheme, which is developed on the basis of the concept of the centralized pattern generator, and can be applied to robotic hip exoskeletons [19, 20]. The adaptive oscillators will control the frequency properties, such as the input signal amplitude and frequency. Lenzi et al [21] suggested an assistive controller utilizing an active frequency oscillator for the robotic hip exoskeleton. A specially built adaptive oscillator was suggested by Lee et al [22]-[24] for the measurement of the gait status by the robotic hip exoskeleton. The aforementioned control approaches were usually conducted using a fixed control strategy with a set of fixed control parameters. Thus, to provide optimal assistance to each wearer, the control of exoskeleton needs to be personalized and finely adjusted.

*Research partly supported by National Science Foundation #1563454, #1563921, #1808752 and #1808898. (Corresponding author: He (Helen) Huang; email: hhuang11@ncsu.edu).

X. Tu, M. Li, M. Liu, and H. Huang are with the NCSU/UNC Department of Biomedical Engineering, NC State University, Raleigh, NC, 27695-7115; University of North Carolina at Chapel Hill, Chapel Hill, NC 27599 USA

J. Si is with the Department of Electrical, Computer, and Energy Engineering, Arizona State University, Tempe, AZ, 85281 USA. (email: si@asu.edu).

X. Tu is also with the Department of Mechanical Engineering, Hubei University of Technology, Wuhan, Hubei, 430068 China (e-mail: tuxikai@gmail.com).

Human-in-the-loop optimization for personalization of the robotic assistance has been an emerging topic recently, but it lacks the ability to adapt to new environments. Gopinath et al [25] proposed a mathematical framework to solve user-driven customization of shared autonomy in assistive robotics with nonlinear optimization techniques. They based their approach on an interactive optimization procedure that customizes control sharing using an assistive robotic arm. Kim et al [26] used a sample-efficient, noise-tolerant, and global optimization approach to minimize the metabolic cost of the walking assistance task by optimizing walking step frequencies. Compared to an existing approach based on gradient descent, this method realized the objectives with lower overall energy expenditure, faster convergence, and smaller inter-subject variability. Zhang et al [27] used a covariance matrix adaptation evolution strategy (CMA-ES) optimization approach to improve the effectiveness of assistive devices by minimizing human energy cost, in order to identify a good assistance pattern and customize it to individual needs. Ding et al [5] used the Bayesian optimization approach and metabolic rate as an optimization index to find optimal peak and offset timing of hip extension assistance by minimizing walking energy costs. Their results indicated that personalized control strategies were better than fixed ones in terms of reducing metabolic cost. Although these methods can customize the control strategies or parameters, they are time-consuming and lack of adaptation. A small change of the wearer requires a re-design of the control.

Reinforcement learning (RL) control is promising technology for optimal adaptive control of wearable robotics as demonstrated in [28-31]. However, these RL control strategies were model-dependent. Vogt et al [28] used learned interaction models to capture the important spatial relationships and temporal synchrony of body movements between two interacting partners. Peng [29] described the dynamics of the lower limb exoskeleton system as a general nonlinear mechanical system and first validated it in a simulation environment before human-related experiments. Hamaya et al [30, 31] exploited a data-efficient model-based reinforcement learning framework called Probabilistic Inference for Learning Control (PILCO) to learn a proper assistive strategy. Generating a well-validated model for the controller generally requires large amount of data, and it can be time-consuming. Additionally, the uncertainties introduced by the inevitable modeling errors can adversely affect control performance.

The goal of this study is to develop a new RL-based optimal adaptive control solution framework for a hip exoskeleton to produce personalized torque assistance over time in order to reduce human efforts during walking. Even though we have developed RL control approaches, specifically data-driven without independent modeling procedures, in our previous studies [32]-[35], they exclusively focus on tackling the parameter personalization problems for robotic knee prostheses.

Our contributions to the new solution framework include the following. (1) Instead of using metabolic cost reduction as the optimization goal, we used proposed transferred work ratio between the exoskeleton and wearer. This ratio can be estimated at a higher rate than metabolic cost by the torque sensor in the exoskeleton without additional instruments such as a mask and a cardiopulmonary system. This is time efficient and user friendly. (2) We designed and implemented a RL controller based on least square policy iteration (LSPI), an approximate policy iteration algorithm to configure the optimal assistive torque profile represented by two timing parameters. (3) We validated this new solution framework on a human wearer and demonstrated its feasibility for personalizing exoskeleton control while walking with the device quickly and effectively.

In the rest of the paper, Section II introduces the new RL based solution framework based on our newly developed 2-DOF robotic hip exoskeleton; Section III introduces human in the loop optimization experiments using LSPI; Section IV presents experimental results and detailed data analyses. Conclusions and future work are presented in Section V.

## II. NEWLY DEVELOPED HIP EXOSKELETON AND CONTROL

### A. Newly Developed Modular Joint 2-DOF Hip Exoskeleton

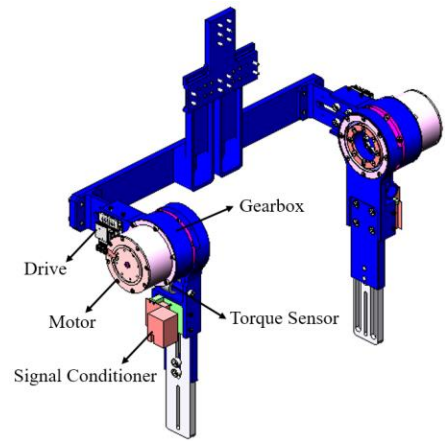

Fig. 1. The mechanical design description of modular joint

Compared with our previous designs [12]-[14], we have recently developed a new lightweight, compact, compliant and fully powered 2-DOF robotic hip exoskeleton with modular joints shown in Fig.1, which can be easily extended with modular design. Unlike the previous designs relying on SEAs, we utilized the admittance control [18] in this new design to realize the equivalent compliance thereby reducing the weight of modular joint from 1.5 kg to 1 kg, while providing higher bandwidth, and a broader range of adjustment for stiffness and damping.

A flat brushless DC motor (UTO-52, Celera Motion, MA, USA) equipped with a high precision incremental encoder (163840 ppr, CE400, Celera Motion, MA, USA), producing a continuous torque rating of 0.353 N·m. The transmission ratio of the harmonic gearbox (CSD-20-100-2A-GR-SP674, Harmonic Drive, MA, USA) is 100:1, resulting in an output torque of approximately 35.3N·m, and a peak torque of approximately 105.9 N·m. Each joint weighs only 1 kg. The entire assembly of the actuator and the transmission is designed to weight 1 kg and it can provide a joint velocity up to 300 deg/sec ,which is sufficient to support fast walking and slow running activities. We use the world's smallest platinum solo twitter (G-SOLTWIR50/100EE1H, Elmo, Israel) as the motor drive with a maximum power of 1000W.

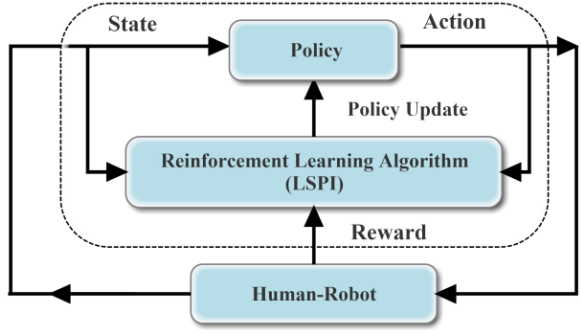

(a)

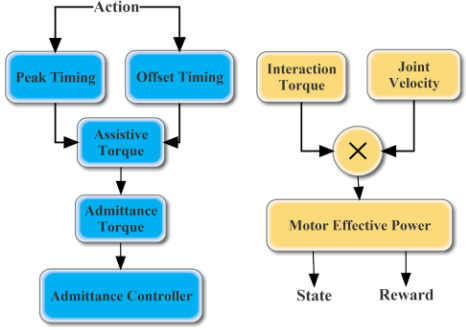

(b)

Fig. 2. Reinforcement learning control, (a). LSPI diagram, (b). The concept definitions of action, state and reward.

### B. Reinforcement Learning Control

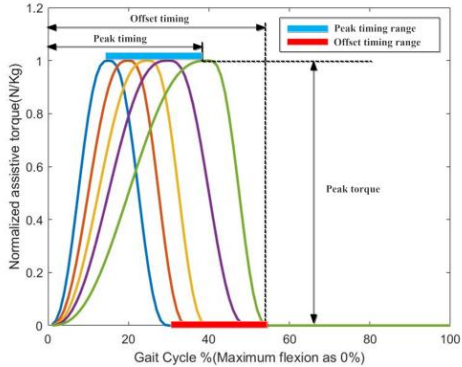

Fig. 3. Assistive torque as a function of peak timing and offset timing

The hip assistive torque as admittance torque was input to admittance controller shown in Fig.2 (b). The admittance controller was executed using the EtherCat protocol based on the TwinCat 3.1 software (Beckhoff Automation, Germany) with sampling rate of 1000Hz. The assistive torque profile was a mixture of two halves of minimum jerk curves joined together at their tops shown in Fig.3. This profile has been specified by two free parameters (peak force and onset timing) that were tuned by reinforcement learning. We set the peak torque to 30% body weight to ensure sufficient assistance while also considering the comfort of the user during walking, which is further confirmed by the prior research [4]. The timing of onset was determined by the maximum hip flexion based on a previous study [5]. The peak and offset timing were limited to 15 to 40% and 30 to 55%, respectively, in this newly established gait cycle with maximum hip flexion as the start time [5].The RL action vector $u$ consisted of adjustable peak and offset timing. The initial timing control parameters $I$ were defined in (1),

$$I = [t_{peak}, t_{offset}]^T \in \Re^2 \quad (1)$$

$$u = \Delta I \quad (2)$$

where $u$ was the action vector and the increment adjustment $\Delta I$ of $I$ in (2), and $t_{peak}$ and $t_{offset}$ were the predefined initial timing control values, respectively.

The interaction behavior between the human and robotic exoskeleton is a process of energy transferring. In order to produce movement assistance, this study designed an exoskeleton applied torque that can maximize the motor effective positive power and minimize the motor effective negative power [6, 39, 40]. Power is the product of measured torque and velocity, and work is the time-based integration of power during a gait cycle. This process is equivalent to maximizing the transferred positive work $W_{motor}^+$ and minimizing the transferred negative work $W_{motor}^-$ in a gait cycle [6, 39, 40].

We define a state $x \in \Re$ in (3) as shown below for the reinforcement learning controller,

$$x = \alpha(\eta - \eta_t) \quad (3)$$

where $\alpha$ is a scaling factor, $\eta$ is the measured transferred work ratio defined as

$$\eta = \frac{|W_{motor}^-|}{W_{motor}^+ + |W_{motor}^-|} \quad (4)$$

and $\eta_t$ in (3) is an empirical target transferred work ratio corresponding to ideal assistances.

In this paper, the human-robot system is taken as a discrete nonlinear system,

$$x_{k+1} = F(x_k, u_k), k = 0, 1, 2, ... \quad (5)$$
$$u_k = \pi(x_k) \quad (6)$$

where $k$ is the discrete time index that represented each timing control parameter update, $x_k \in \Re$ is the state vector $x$ at k-th update, $u_k \in \Re^2$ is the action vector $u$ at k-th update, $F$ is the unknown system dynamics, and $\pi: \Re \to \Re^2$ is the control policy. To provide the adaptive learning control of the human-robot interaction system, we define a stage cost with the following quadratic form for RL controller,

$$U(x, u) = x^T R_x x + u^T R_u u \quad (7)$$

where $R_x \in \Re^{1 \times 1}$ and $R_u \in \Re^{2 \times 2}$ are positive definite matrices. We use (12) to regulate state $x$ and action $u$, as larger state deviation as in (3) and larger action adjustment as in (2) will be penalized with a larger cost. LSPI RL [36]-[38] was used to minimize the reward of the equation (7). The quadratic basis function was used to calculate the cost function. As there were one 1-D state and one 2-D action in this study, the RL policy learnable weight vector was 6-D. A sample consisted of state, action, reward and next state. The number of samples was usually selected more than twice the weight vector dimension. The derived process of the online LSPI RL optimal control was referred to the studies [37, 38]. The reinforcement learning controller shown in Fig. 2 was realized in MATLAB, which communicated with the admittance controller by the real-time UDP.

### III. HUMAN IN THE LOOP OPTIMIZATION EXPERIMENTS

#### A. Experimental Protocol and Setup

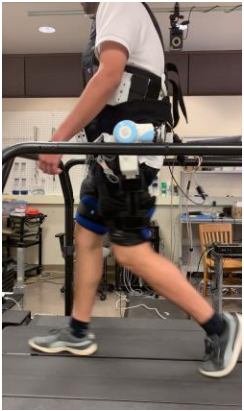

Fig. 4. The subject walking with the hip exoskeleton in the experiment

The experimental protocol was approved by the Institutional Review Board (IRB) of University of North Carolina at Chapel Hill. The proposed controller was experimentally tested with a male able-bodied subject (80kg of weight, 172 cm of height; 36 years of age). He had no history of lower limb injury or neurodegenerative disorder. The subject practiced walking with hip extension assistance on a day prior to the experiment to ensure that he was comfortable in walking with the exoskeleton. The surface EMG electrode was placed on the gluteus maximus of the subject's left hip. The ground electrode was placed on the radial styloid of right wrist. EMG data were collected by MA400-28 EMG system (Motion Lab Systems, Inc., Baton Rouge, LA, USA) and sampled at a rate of 1000 Hz. The subject walked on a dual-belt treadmill (1000 Hz, Bertec Corp., Columbus, OH, USA) at a fixed speed of 1.3 m/s under three different modes.

Free Walking: The subject walked for two minutes without wearing the hip exoskeleton in order to measure the baseline EMG activity of the gluteus maximus during walking.

Zero Torque: The subject walked for two minutes with the exoskeleton attached and controlled in a transparent mode. In this mode, the peak torque of hip assistance was set to zero hence operating in "Zero Torque" mode. The hip exoskeleton admittance controller parameters were set to minimize the interaction torque with the wearer's leg in order to reduce the loading effect of the robot. This mode was used as a basis for the evaluation of assistance mode to understand the impact of the exoskeleton assistance on transferred power and energy in addition to muscle recruitment during optimal assistance.

Assisted Torque: In order to validate the proposed RL control, hip extension assistance was provided to the left side while the right hip robotic joint operated in the zero-torque mode. The assistive torque was provided to the wearer after he walked in the zero-torque mode for two minutes. The initial control parameters: peak timing and offset timing were randomly selected in this mode. The assistive torque was gradually increased from zero to the target peak torque in a span of 15 steps.

To reduce the influence of noises introduced by the inherent human variance during walking, a window of length 5 gait cycles or steps were set as a RL action update period (i.e., the time index in (5)).The state was obtained with an average of every 5 gait cycles. The proposed online LSPI RL controller was used to automatically update the action: two timing control parameters according to the state-action policy function. A policy was updated every 15 samples (75 gait cycles) if the stopping criterion was not met. An acceptable error range was considered as the RL stopping criterion due to the human variance in walking. If the reward errors stayed within the target set for 10 consecutive action updates, the assisted-torque mode would be considered reaching the stopping criterion. In this study, the bound range of reward errors was set to 0.01. In order to prevent subject fatigue, the training would be stopped automatically if the assisted-torque mode training time exceeded 15 minutes. The assistive mode was switched to zero-torque mode if RL control met the stopping criterion. If an assisted trial with a well-learned policy took less than 15 minutes, it was regarded as a successful trial. Four assisted experimental trials were conducted. The last 20 steps of assisted-torque mode in a successful trial were considered as an "optimal assistance" phase. The maximum flexion as the start of gait cycle was detected by the hip joint position sensor. The target transferred work ratio in (3) was set to 0.16 based on experience and the scaling factor was 7. The subject wore a 2-DOF hip exoskeleton and walked on the treadmill as shown in Fig. 4.

#### B. Evaluation Measures

Measures of transferred power and work during a gait cycle were obtained with an average of the last 20 strides in zero-torque and assisted-torque walking modes. Transferred power was compared among the "optimal assistance" phases of successful trials and the zero-torque phase of the first trial. The transferred positive and negative work were calculated separately by integrating the area under the power curve during an averaged gait cycle. For processing the EMG signals, we band-pass filtered (Butterworth, 20-300 Hz, 4th order), full-wave rectified, and low-pass filtered (Butterworth, 1 Hz, 4th order) them to create a linear envelope. The envelope was then averaged to a gait cycle with last 20 strides in three different modes. EMG data were normalized by the maximum peak observed during the last 20 strides of free walking. All gait events were identified by ground reaction forces. EMG activities of the same muscle were compared in a gait cycle with the last 20 strides averaged among the "optimal assistance" phases of successful trials, the free-walking base line and the zero-torque phase of the first trial. The root mean square (RMS) of EMG and standard derivation were calculated, in order to evaluate the human hip extension effort.

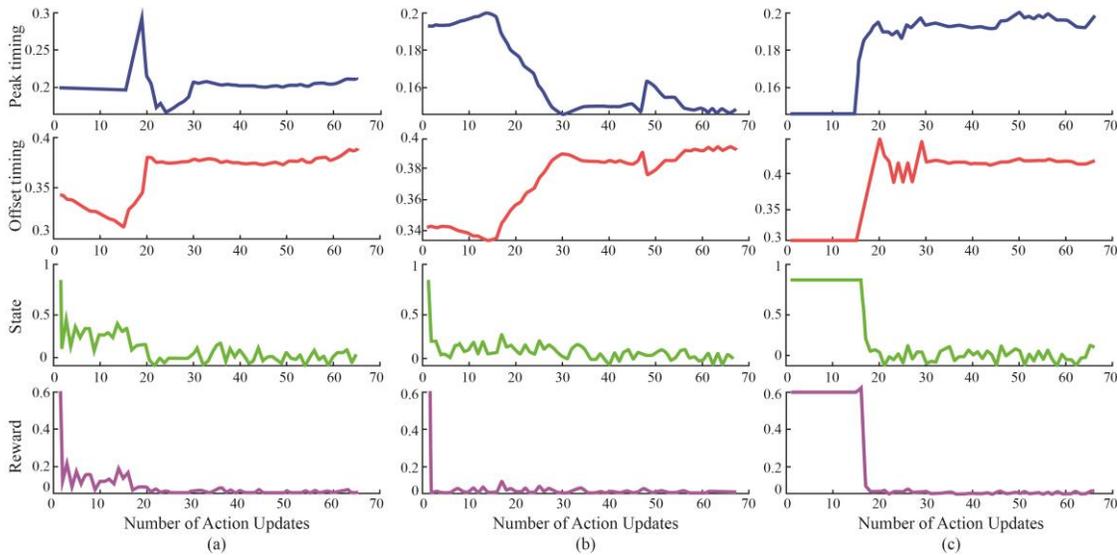

Fig. 5. Evolution of peak timing, offset timing, state and reward in three successful assisted-torque trials. (a) Trial 1 , (b) Trial 2, (c) Trial 3.

## IV. EXPERIMENTAL RESULTS

The recruited subject participated in four assistive torque trials. Wherein, three trials were successful and one stopped due to overtime. These "optimal assistance" periods in the three successful trials were called phases of Optimal Assistance 1, 2 and 3, respectively.

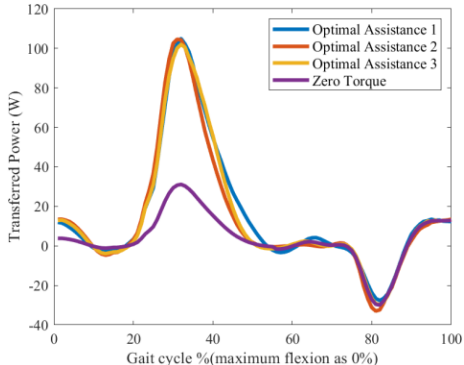

Fig. 6. Transferred power in the human-robot interaction

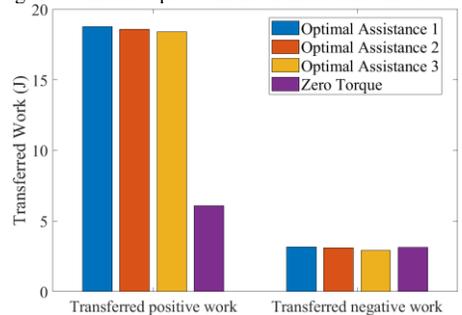

Fig. 7. Transferred work in an averaged gait cycle

### A. Evaluation on RL successful trials

The total number of action updates in these three trials is 65, 67 and 66, respectively. The policy was updated thrice for each of the successful trials, executed at the action update number 15, 30 and 45, respectively. As Fig. 5 shows, the states and rewards in the initial stages were distant from the target ones in Trial 1. Clearly, after the policy was updated once, the control parameters: peak and offset timing were both adjusted to obvious larger values by the proposed RL controller, which made the states and rewards gradually approach the targets. The peak timing and offset timing for optimal assistance during Trial 1 were found to be 22.67% and 38.86%. The peak and offset timing during Trial 2 was unchanged before the policy was updated. Clearly, after the policy was first updated, peak timing decreased while offset timing increased. The optimal assistance was achieved for Trial 2 with peak timing 15.27% and offset timing 39.39%. In Trial 3, the states and rewards before the first policy update were always kept at the initial values, which were caused by peak or offset timing going out of the action bound along with each action update. From the start of the first updated policy, peak and offset timing, state, and reward suddenly changed. Parameters for optimal assistance in Trial 3 were found: peak timing 20.78% and offset timing 41.28%. The results of three trials indicated that LSPI RL controller was able to realize the optimal assistance from different initial parameters. The similarities of control parameters in these three optimal assistance phases were found. The assisted-toque training time of these three successful trials was 347, 369 and 358 seconds, respectively, which were much less than the average time 1284 seconds required for optimization using the Bayesian approach [5] indicating that LSPI RL method was more time-efficient.

### B. Transferred power and energy from exoskeleton to human

Transferred power and work measures provided a quantitative evaluation on the performance of RL-based control. Fig. 6 illustrates the effective motor power supplied by the hip exoskeleton to the wearer in three "optimal assistance" phases and the zero-torque mode. The positive power amplitude peaks with three optimal assistance phases were 101.2 W, 101.1 W and 98.51W, respectively, while the one in the zero-torque mode was 31.12W. The optimal assistance phases had much higher peak positive power compared to the zero torque condition, while the peak negative power was similar across trials.  These results showed that RL control can maximize the transferred positive power to the wearer compared with the zero-torque mode. The total transferred positive and negative work across an averaged gait cycle is shown in Fig. 7 where it can be inferred that there was no statistical difference in the negative work across all trials. While there was no significant difference in positive wok across the three optimal assistance phases, all the three of them were significantly higher than the zero-torque mode. By comparing the measured transferred work ratio with the target ratio, the performance of the RL controller was assessed. The target transferred work ratios for three "optimal assistance" phases were 0.1593, 0.1621 and 0.1602, respectively. As the zero-torque mode was realized by the minimization of the interaction torque between the wearer and the hip exoskeleton, the ratio with zero-torque mode was 0.3396, which indicates that some positive work was still transferred even under the zero-torque condition. The behavior of the transferred work in these three trials with optimal assistance was consistent both temporally and quantitatively validating the feasibility of the proposed LSPI RL controller. The analysis reflected how the transferred work ratio shifted with the exoskeleton supplying more positive work and thereby reduced human efforts.

*C. Muscle effort reduction*

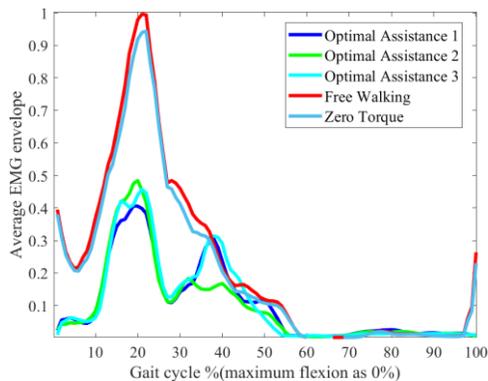

Fig. 8. EMG activities of gluteus maximus with different conditions within one averaged gait cycle

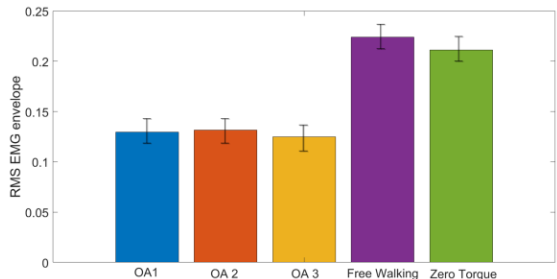

Fig. 9. RMS EMG during an averaged gait cycle under different conditions, OA: optimal assistance.

The linear envelope of the recorded EMG signals was computed to generate the temporal pattern of the muscle activation shown in Fig.8, which represents EMG activity of the hip extensor gluteus maximus under the five conditions: three "optimal assistance" phases, free-walking and zero-torque. The peak of the linear envelope shows the maximum activation, which is well correlated to the maximum force produced by this muscle. The averaged EMG peaks in sequence are 0.4064, 0.4841, 0.4546, 0.996 and 0.9437, respectively. The peaks of three optimal assistance phases were reduced to 59.24%, 51.40% and 54.36% of the EMG peak amplitude during free walking. Fig. 9 shows the RMS EMG envelope as an indication of the total effort spent during a gait cycle during walking. The mean values for these five conditions are 0.1297, 0.1317, 0.1249, 0.2241 and 0.2113, respectively. The mean value in the zero-torque mode was a slightly lower than that of free-walking mode in Fig. 9, which indicated that there was some transferred positive energy to the wearer in the zero-torque mode. The results of EMG activities demonstrated that LSPI RL controller can reduce the muscle contraction intensity with the optimal assistance.

The timing control parameters exhibited somewhat oscillatory patterns between the first and the second policy update of the second successful trial shown in Fig. 5. This was because the RL takes trial and error strategies, involving the dynamical interactions between the wearer and the hip exoskeleton. Such varying interactions would introduce more perturbations to the robot and resulted in oscillations. Under the above discussed disturbances, the RL controller responded by adjustment when it observed discrepancies between target and actual states. The reason for one trial failure was that one initial RL policy was randomly generated, especially one nonideal policy, which cannot guarantee a successful trial with less than 15 minutes. These unique phenomena are the results of dealing with an inherently co-adapting human-robot RL-based control system.

## V. CONCLUSION AND FUTURE WORK

In this paper, a novel LSPI RL adaptive optimal controller was developed to configure timing parameters of assistive torque automatically for a robotic hip exoskeleton. The online learning controller was able to successfully learn the control policy as was demonstrated not only by increased the transferred positive work, but also by the decreased EMG activity of the hip extensor muscle. These results indicated that that the online LSPI RL controller is a promising new tool to solve the challenging parameter tuning problems for personalized walking assistance.

We only applied this newly designed reinforcement learning to hip extension assistance and validated the method on one human subject. Further studies need to be done to investigate whether the outcome of the proposed method can be adapted for both flexion and extension assistance of both the hip joints in coordination. Studies using additional subjects can further verify if the proposed assistance protocol can be generalized across subjects. In addition, our future work will extend the current design to facilitate further online control policy adjustment. We believe such an integrated approach will facilitate even broader range of the human-robot interaction behavior to address changes in environment, task, and human physical condition.

ACKNOWLEDGE

We thank Sameer Amarnath Upadhye, Varun Nalam, Rajat Emanuel Singh, Abbas Alili and Chinmay Shah for their assistance in the experiments and Varun Nalam for editing the paper.

REFERENCES

[1] B. Chen, B. Zi, L. Qin, and Q. Pan, "State-of-the-art research in robotic hip exoskeletons: A general review," Journal of Orthopaedic Translation, vol. 20, pp. 4-13, 2020.
[2] D. J. Farris and G. S. Sawicki, "The mechanics and energetics of human walking and running: a joint level perspective," Journal of The Royal Society Interface, p. rsif20110182, 2011
[3] G. S. Sawicki, C. L. Lewis, and D. P. Ferris, "It pays to have a spring in your step," Exercise and sport sciences reviews, vol. 37, no. 3, pp. 130, 2009.

[4] Y. Ding, F. A. Panizzolo, C. Siviy, P. Malcolm, I. Galiana, K. G. Holt, and C. J. Walsh, "Effect of timing of hip extension assistance during loaded walking with a soft exosuit," Journal of neuroengineering and rehabilitation, vol. 13, no. 1, pp. 1-10, 2016.

[5] Y. Ding, M. Kim, S. Kuindersma, and C. J. Walsh, "Human-in-the-loop optimization of hip assistance with a soft exosuit during walking," Science Robotics, vol. 3, no. 15, pp. eaar5438, 2018.

[6] B. Lim, J. Lee, J. Jang, K. Kim, Y. J. Park, K. Seo, and Y. Shim, "Delayed output feedback control for gait assistance with a robotic hip exoskeleton," IEEE Transactions on Robotics, vol. 35, no. 4, pp. 1055-1062, 2019.

[7] F. A. Panizzolo, I. Galiana, A. T. Asbeck, C. Siviy, K. Schmidt, K. G. Holt, and C. J. Walsh, "A biologically-inspired multi-joint soft exosuit that can reduce the energy cost of loaded walking," Journal of neuroengineering and rehabilitation, vol. 13, no. 1, pp. 1-14, 2016.

[8] I. Kang, H. Hsu, and A. Young, "The effect of hip assistance levels on human energetic cost using robotic hip exoskeletons," IEEE Robotics and Automation Letters, vol. 4, no. 2, pp. 430-437, 2019.

[9] F. Giovacchini, F. Vannetti, M. Fantozzi, M. Cempini, M. Cortese, A. Parri, T. Yan, D. Lefeber, and N. Vitiello, "A light-weight active orthosis for hip movement assistance," Robotics and Autonomous Systems, vol. 73, pp. 123-134, 2015.

[10] B. Chen, L. Grazi, F. Lanotte, N. Vitiello, and S. Crea, "A real-time lift detection strategy for a hip exoskeleton," Frontiers in neurorobotics, vol. 12, pp. 17, 2018.

[11] J. Masood, J. Ortiz, J. Fernández, L. A. Mateos, and D. G. Caldwell, "Mechanical design and analysis of light weight hip joint Parallel Elastic Actuator for industrial exoskeleton," 2016 6th IEEE International Conference on Biomedical Robotics and Biomechatronics (BioRob), Singapore, 2016, pp. 631-636.

[12] T. Zhang, M. Tran, and H. Huang, "Design and experimental verification of hip exoskeleton with balance capacities for walking assistance," IEEE/ASME Transactions on Mechatronics, vol. 23, no. 1, pp. 274-285, 2018.

[13] T. Zhang, and H. Huang, "A lower-back robotic exoskeleton: Industrial handling augmentation used to provide spinal support," IEEE Robotics & Automation Magazine, vol. 25, no. 2, pp. 95-106, 2018.

[14] T. Zhang, and H. Huang, "Design and control of a series elastic actuator with clutch for hip exoskeleton for precise assistive magnitude and timing control and improved mechanical safety," IEEE/ASME Transactions on Mechatronics, vol. 24, no. 5, pp. 2215-2226, 2019.

[15] H. Kazerooni, R. Steger, and L. Huang, "Hybrid control of the Berkeley lower extremity exoskeleton (BLEEX)," The International Journal of Robotics Research, vol. 25, no. 5-6, pp. 561-573, 2006.

[16] H. Kawamoto, and Y. Sankai, "Power assist method based on phase sequence and muscle force condition for HAL," Advanced Robotics, vol. 19, no. 7, pp. 717-734, 2005.

[17] A. Martinez, B. Lawson, and M. Goldfarb, "A controller for guiding leg movement during overground walking with a lower limb exoskeleton," IEEE Transactions on Robotics, vol. 34, no. 1, pp. 183-193, 2018.

[18] U. Nagarajan, G. Aguirre-Ollinger, and A. Goswami, "Integral admittance shaping: A unified framework for active exoskeleton control," Robotics and Autonomous Systems, vol. 75, pp. 310-324, 2016.

[19] T. Yan, A. Parri, V. R. Garate, M. Cempini, R. Ronsse, and N. Vitiello, "An oscillator-based smooth real-time estimate of gait phase for wearable robotics," Autonomous Robots, vol. 41, no. 3, pp. 759-774, 2017.

[20] T. G. Sugar, E. Fernandez, D. Kinney, K. W. Hollander, and S. Redkar, "HeSA, hip exoskeleton for superior assistance," Wearable Robotics: Challenges and Trends, pp. 319-323: Springer, 2017.

[21] T. Lenzi, M. C. Carrozza, and S. K. Agrawal, "Powered hip exoskeletons can reduce the user's hip and ankle muscle activations during walking," IEEE Transactions on Neural Systems and Rehabilitation Engineering, vol. 21, no. 6, pp. 938-948, 2013.

[22] H.-J. Lee, S. Lee, W. H. Chang, K. Seo, Y. Shim, B.-O. Choi, G.-H. Ryu, and Y.-H. Kim, "A wearable hip assist robot can improve gait function and cardiopulmonary metabolic efficiency in elderly adults," IEEE transactions on neural systems and rehabilitation engineering, vol. 25, no. 9, pp. 1549-1557, 2017.

[23] K. Seo, J. Lee, Y. Lee, T. Ha and Y. Shim, "Fully autonomous hip exoskeleton saves metabolic cost of walking," 2016 IEEE International Conference on Robotics and Automation (ICRA), Stockholm, 2016, pp. 4628-4635

[24] K. Seo, K. Kim, Y. J. Park, J.-K. Cho, J. Lee, B. Choi, B. Lim, Y. Lee, and Y. Shim, "Adaptive oscillator-based control for active lower-limb exoskeleton and its metabolic impact." IEEE International Conference on Robotics and Automation (ICRA), Brisbane, QLD, 2018, pp. 6752-6758.

[25] D. Gopinath, S. Jain, and B. D. Argall, "Human-in-the-loop optimization of shared autonomy in assistive robotics," IEEE Robotics and Automation Letters, vol. 2, no. 1, pp. 247-254, 2016.

[26] M. Kim, Y. Ding, P. Malcolm, J. Speeckaert, C. J. Siviy, C. J. Walsh, and S. Kuindersma, "Human-in-the-loop Bayesian optimization of wearable device parameters," PloS one, vol. 12, no. 9, pp. e0184054, 2017.

[27] J. Zhang, P. Fiers, K. A. Witte, R. W. Jackson, K. L. Poggensee, C. G. Atkeson, and S. H. Collins, "Human-in-the-loop optimization of exoskeleton assistance during walking," Science, vol. 356, no. 6344, pp. 1280-1284, 2017.

[28] D. Vogt, S. Stepputtis, S. Grehl, B. Jung, and H. Ben Amor, "A system for learning continuous human-robot interactions from human-human demonstrations," in Proc. - IEEE Int. Conf. Robot. Autom., 2017.

[29] Z. Peng, R. Luo, R. Huang, J. Hu, K. Shi, H. Cheng, and B. K. Ghosh, "Data-Driven Reinforcement Learning for Walking Assistance Control of a Lower Limb Exoskeleton with Hemiplegic Patients," 2020 IEEE International Conference on Robotics and Automation (ICRA), Paris, France, 2020, pp. 9065-9071

[30] M. Hamaya, T. Matsubara, T. Noda, T. Teramae, and J. Morimoto, "Learning assistive strategies for exoskeleton robots from user-robot physical interaction," Pattern Recognition Letters, vol. 99, pp. 67-76, 2017.

[31] M. Hamaya, T. Matsubara, T. Noda, T. Teramae and J. Morimoto, "Learning task-parametrized assistive strategies for exoskeleton robots by multi-task reinforcement learning," 2017 IEEE International Conference on Robotics and Automation (ICRA), Singapore, 2017, pp. 5907-5912

[32] Y. Wen, J. Si, X. Gao, S. Huang, and H. H. Huang, "A new powered lower limb prosthesis control framework based on adaptive dynamic programming," IEEE Transactions on Neural Networks and Learning Systems, vol. 28, no. 9, pp. 2215–2220, 2017

[33] Y. Wen, J. Si, A. Brandt, X. Gao and H. H. Huang, "Online Reinforcement Learning Control for the Personalization of a Robotic Knee Prosthesis," IEEE Transactions on Cybernetics, vol. 50, no. 6, pp. 2346-2356, June 2020

[34] Y. Wen, M. Li, J. Si and H. Huang, "Wearer-Prosthesis Interaction for Symmetrical Gait: A Study Enabled by Reinforcement Learning Prosthesis Control," IEEE Transactions on Neural Systems and Rehabilitation Engineering, vol. 28, no. 4, pp. 904-913, April 2020

[35] M. Li, X. Gao, Y. Wen, J. Si, and H. Huang, "Offline policy iteration based reinforcement learning controller for online robotic knee prosthesis parameter tuning," in 2019 IEEE International Conference on Robotics and Automation (ICRA), May 2019, pp. 2831–2837

[36] J. Si and Yu-Tsung Wang, "Online learning control by association and reinforcement," in IEEE Transactions on Neural Networks, vol. 12, no. 2, pp. 264–276, 2001.

[37] L. Buşoniu, D. Ernst, B. De Schutter, and R. Babuška, "Online least-squares policy iteration for reinforcement learning control." Proceedings of the American Control Conference, pp. 486-491, 2010.

[38] M. G. Lagoudakis, and R. Parr, "Least-squares policy iteration," Journal of machine learning research, vol. 4, no. Dec, pp. 1107-1149, 2003.

[39] L. M. Mooney, E. J. Rouse, and H. M. Herr, "Autonomous exoskeleton reduces metabolic cost of human walking during load carriage," Journal of neuroengineering and rehabilitation, vol. 11, no. 1, pp. 1-11, 2014.

[40] M. B. Yandell, B. T. Quinlivan, D. Popov, C. Walsh, and K. E. Zelik, "Physical interface dynamics alter how robotic exosuits augment human movement: implications for optimizing wearable assistive devices," Journal of neuroengineering and rehabilitation, vol. 14, no. 1, pp. 40, 2017